\documentclass[letterpaper]{article} 
\usepackage[submission]{aaai2026}  
\usepackage{times}  
\usepackage{helvet}  
\usepackage{courier}  
\usepackage[hyphens]{url}  
\usepackage{graphicx} 
\usepackage{natbib}  
\usepackage{caption} 
\usepackage{algorithm}
\usepackage{algorithmic}
\usepackage{amsmath,amssymb,amsfonts}
\usepackage{booktabs}
\usepackage{adjustbox}
\usepackage{multirow}
\usepackage{newfloat}
\usepackage{listings}
\DeclareCaptionStyle{ruled}{labelfont=normalfont,labelsep=colon,strut=off} 
\floatstyle{ruled}
\newfloat{listing}{tb}{lst}{}
\floatname{listing}{Listing}
\title{TNet: Terrace Convolutional Decoder Network for Remote Sensing Image Semantic Segmentation}
\author {
    Chengqian Dai\textsuperscript{\rm 1},
    Yonghong Guo\textsuperscript{\rm 1},
    Zhaohong Xiang\textsuperscript{\rm 1},
    Yigui Luo\textsuperscript{\rm 1},
}
\affiliations {
    \textsuperscript{\rm 1}Tongji University\\
}

\begin{document}

\maketitle

\begin{abstract}
In remote sensing, most segmentation networks adopt the UNet architecture, often incorporating modules such as Transformers or Mamba to enhance global-local feature interactions within decoder stages. However, these enhancements typically focus on intra-scale relationships and neglect the global contextual dependencies across multiple resolutions.
To address this limitation, we introduce the Terrace Convolutional Decoder Network (TNet), a simple yet effective architecture that leverages only convolution and addition operations to progressively integrate low-resolution features (rich in global context) into higher-resolution features (rich in local details) across decoding stages. This progressive fusion enables the model to learn spatially-aware convolutional kernels that naturally blend global and local information in a stage-wise manner.
We implement TNet with a ResNet-18 encoder (TNet-R) and evaluate it on three benchmark datasets. TNet-R achieves competitive performance with a mean Intersection-over-Union (mIoU) of 85.35\% on ISPRS Vaihingen, 87.05\% on ISPRS Potsdam, and 52.19\% on LoveDA, while maintaining high computational efficiency. Code is publicly available.
\end{abstract}

\section{Introduction}\label{Intro}

Semantic segmentation of high-resolution remote sensing imagery plays a vital role in applications such as urban planning~\cite{zhang2024uv}, environmental monitoring~\cite{himeur2022using}, and disaster assessment~\cite{fu2023cal}. It involves pixel-wise classification of image content, enabling fine-grained recognition and annotation of diverse land cover types and objects. However, targets in remote sensing images are often small and densely distributed, making it difficult to rely solely on local details for accurate identification. Instead, effective segmentation requires the integration of both \textbf{global contextual cues} and \textbf{local spatial details} to distinguish objects based on holistic semantics and appearance.

To address this, numerous methods based on the UNet architecture have been proposed to enhance global-local feature fusion. Recent works such as BANet~\cite{wang2021transformer}, SFANet~\cite{hwang2024sfa}, UNetFormer~\cite{wang2022unetformer}, DC-Swin~\cite{9681903}, CMTFNet~\cite{10247595}, and UMFormer~\cite{10969832} incorporate Transformer-based attention mechanisms~\cite{vaswani2017attention, liu2021swin, dosovitskiy2020image} into convolutional frameworks to enrich contextual understanding at various resolutions. Others, including RS3-Mamba~\cite{10556777}, CMUNet~\cite{liu2024cm}, and PyramidMamba~\cite{wang2024pyramidmamba}, adopt state-space models such as Mamba~\cite{gu2023mamba, zhu2024vision} to better capture long-range dependencies.


\begin{figure}[t]
\centering
\includegraphics[width=1\columnwidth]{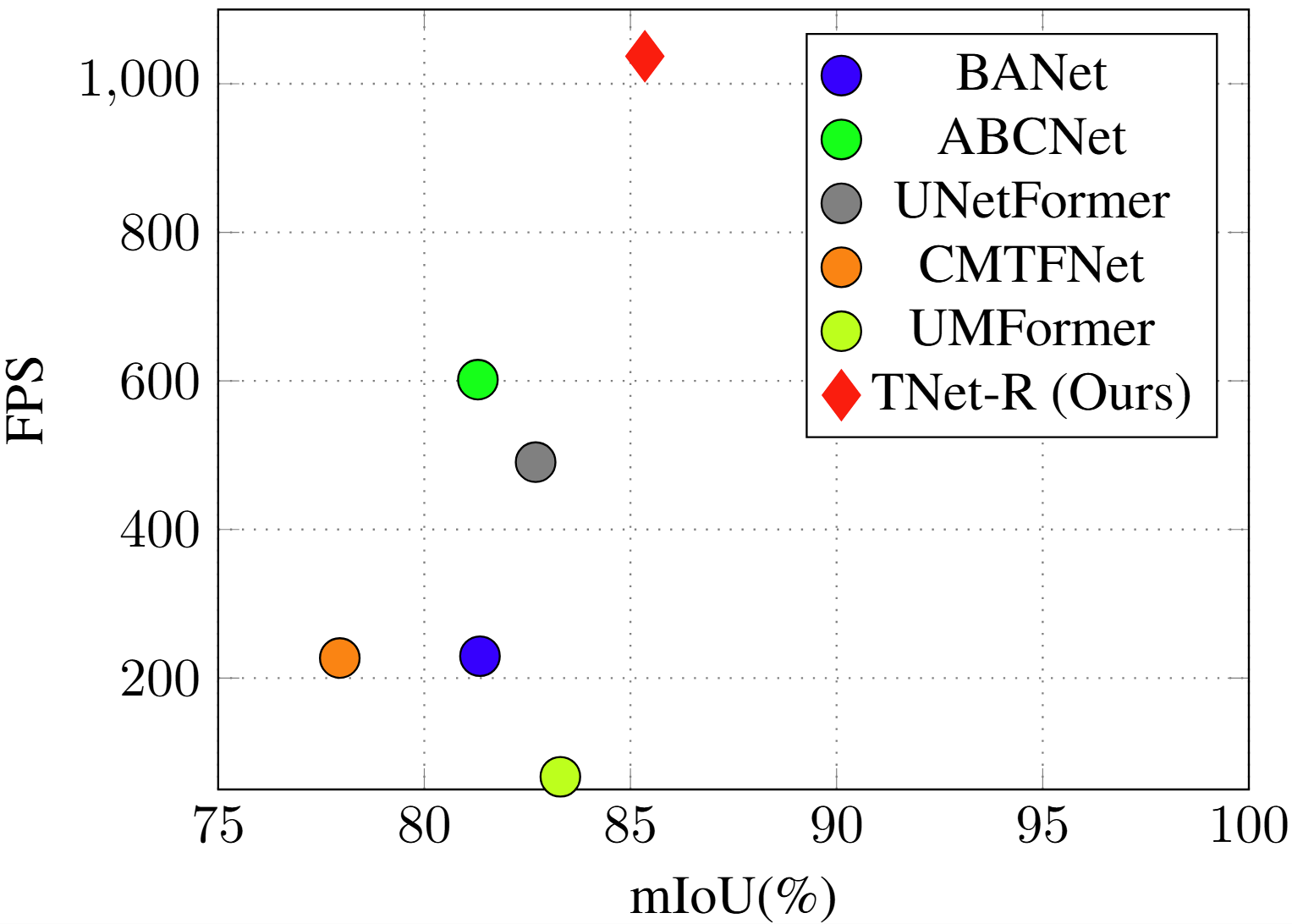} 
\caption{The comparison of mIoU and FPS is conducted with an input image size of $3\times512\times512$. The mIoU results are derived from the ISPRS Vaihingen dataset, and the inference is performed on an Nvidia RTX 4090. Our method achieves the best performance.}
\label{fig:plot}
\end{figure}


Meanwhile, recent studies have focused on developing more efficient decoder architectures tailored for the remote sensing domain. For example, DANet~\cite{fu2019dual} employs dual branches to capture features, effectively integrating both channel and positional information. D2LS~\cite{d2ls} proposes a dynamic dictionary learning framework that explicitly models class-aware semantic embeddings and uses a contrastive loss to improve intra-class compactness and inter-class separability. Similarly, Mask2Former~\cite{cheng2022masked} introduces masked attention in the Transformer decoder, limiting attention to localized features around predicted segments. These studies emphasize the importance of designing efficient, task-specific decoders that balance global context integration with the preservation of fine-grained local details.


Motivated by these observations, we propose the \textbf{Terrace Convolutional Decoder Network (TNet)}, a simple yet effective decoder that progressively fuses global semantics and local details through a terrace-like structure. TNet relies only on standard convolution and addition operations, enabling efficient and scalable cross-scale fusion. Built atop a generic encoder (e.g., ResNet~\cite{he2016deep}), TNet progressively upsamples and integrates multi-level features via transposed convolutions to reconstruct high-resolution predictions. We instantiate TNet with ResNet18 (TNet-R) and evaluate it on the ISPRS Vaihingen, ISPRS Potsdam~\cite{ISPRS_Vaihingen_and_ISPRS_Potsdam_2024}, and LoveDA~\cite{junjue_wang_2021_5706578} datasets, achieving state-of-the-art performance with low computational complexity.

The contributions of this paper are as follows:
\begin{itemize}
    \item We propose \textbf{TNet}, a terrace-shaped convolutional decoder that progressively integrates global context into local details using only standard convolutions, achieving strong performance with minimal complexity.
    \item We demonstrate the \textbf{modularity and flexibility} of TNet by applying it to different encoders, showing its compatibility with a variety of backbone architectures.
    \item We show that our ResNet18-based \textbf{TNet-R} achieves state-of-the-art segmentation results on three public benchmarks, validating the effectiveness of our design.
\end{itemize}

\section{Related Work}

\subsection{Remote Sensing Image Semantic Segmentation}
Remote sensing image semantic segmentation is challenging because targets are often small and easily overlooked. 
This necessitates methods that capture both global context and fine local details.
CNNs remain a foundational approach for segmentation; For instance, DANet~\cite{fu2019dual} employs a dual-branch mechanism to integrate positional and channel attention features. 
Similarly, ABCNet~\cite{li2021abcnet} and BANet~\cite{wang2021transformer} extend the UNet architecture by adding a parallel pathway to better capture global context.

With the rise of Transformers in vision, many models leverage self-attention for global context with local details.
Segmenter~\cite{strudel2021segmenter} eliminates CNNs entirely, using a pure Transformer encoder-decoder. 
UNetFormer~\cite{wang2022unetformer} combines CNNs with a global-local attention mechanism to capture context at multiple scales. 
CMTFNet~\cite{10247595} introduces a multiscale multihead self-attention (M2SA) module to extract rich multiscale contextual and channel information. 
Furthermore, other innovations focus on frequency and localization:
AFENet~\cite{10955240} fuses adaptive frequency information with spatial features to better represent fine details, and  Mask2Former~\cite{cheng2022masked} uses a Transformer-based mask attention mechanism to localize features by restricting attention to predicted mask regions. 

Visual State Space Models (e.g., the Mamba module) have been introduced into CNN-based segmentation networks to better unify global and local representations. 
For example, RS3Mamba~\cite{10556777} integrates the Mamba module into the encoder, effectively capturing the relationships between global and local features during the encoding stage. 
On the other hand, CM-Unet~\cite{liu2024cm} focuses on the decoder stage, proposing a Mamba-based framework to efficiently integrate local and global information for improved segmentation performance. 
Finally, UMFormer~\cite{10969832} combines both Transformers
and Mamba modules, aiming to balance efficiency and accuracy.

These methods have effectively improved the performance of image semantic segmentation. From these approaches, we not only summarized the successful strategies of these modules but also observed that the global interactions in the aforementioned modules are almost exclusively confined to single-layer operations.
To address this, we propose the Terrace Convolutional Decoder Network, which ensures that global features from lower-resolution layers are
propagated into higher-resolution layers by progressively adding feature maps across decoder stages.

\subsection{Multi-Scale Feature Fusion}
UNet~\cite{ronneberger2015u} fuses features by upsampling and concatenating higher-resolution encoder features with decoder features, progressively injecting fine detail as resolution increases. FPN~\cite{lin2017feature} instead uses an additive top-down pathway, producing a multi-scale feature pyramid for subsequent processing.
Notably, both approaches ultimately rely on convolution layers to learn the fused feature representations.

For example, SFA-Net~\cite{hwang2024sfa} incorporates multi-scale information into UNet’s skip connections to enhance global feature fusion, and Hi-ResNet~\cite{10638169} similarly improves segmentation via multi-scale feature fusion strategies.
Inspired by these works, we sought to simplify the multi-scale fusion
process. We remove complex operations like feature concatenation and use straightforward addition plus transposed convolution for merging features. This design makes feature propagation more direct and keeps the architecture simple while still effectively integrating
multiple scales.

\begin{figure*}[t]
\centering
\includegraphics[width=1\textwidth]{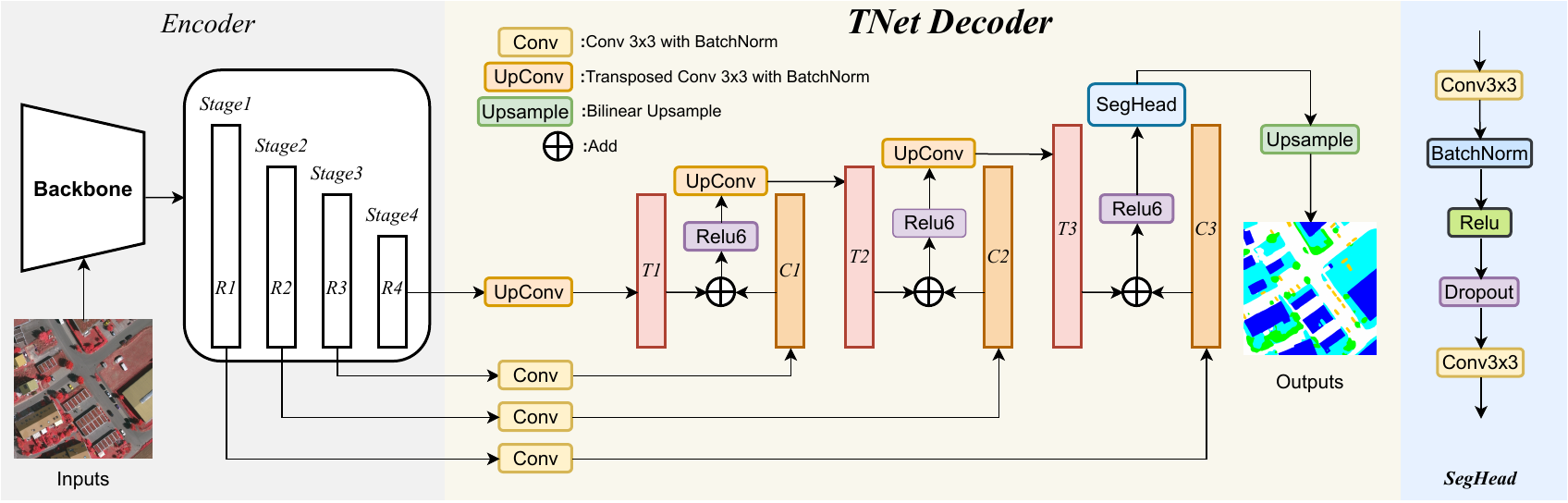} 
\caption{The overview of TNet with the backbone is illustrated in the figure. On the left, the encoder utilizes the backbone to output features from four stages. The input image has a size of $H\times W\times3$, and the features $R1$, $R2$, $R3$, and $R4$ have dimensions of $\frac{H}{4}\times \frac{W}{4}\times C_1$, $\frac{H}{8}\times \frac{W}{8}\times C_2$, $\frac{H}{16}\times \frac{W}{16}\times C_3$ and  $\frac{H}{32}\times \frac{W}{32}\times C_4$, respectively. For example, in ResNet18, $C_1$, $C_2$, $C_3$ and $C_4$ are 64, 128, 256 and 512, respectively. Correspondingly, in the decoder, $T1$'s $C1$ size matches $R3$, $T2$'s $C2$ size matches $R2$, and $T3$'s $C3$ size matches $R1$. Finally, the segmentation head (seg head) outputs the semantic segmentation result, which is upsampled to the original input resolution for the final output.}
\label{fig:overview}
\end{figure*}

\section{Method}
In this section, we describe the architecture of TNet, our proposed decoder. To illustrate TNet in a full segmentation model, we pair it with a ResNet18 encoder (denoted TNet-ResNet or TNet-R) and detail the design and functionality of this configuration.

\subsection{Convolutional Encoder}
TNet can be combined with most backbones to form "TNet-X", such as ResNet~\cite{he2016deep}, ConvNeXt~\cite{liu2022convnet}, EfficientNet~\cite{tan2019efficientnet}, ViT ~\cite{dosovitskiy2020image}, MobileNetV2~\cite{sandler2018mobilenetv2}, and EfficientNetV2~\cite{tan2021efficientnetv2}. In this section, we take the commonly used ResNet18 as an example to provide a detailed explanation of the architecture of TNet-R.

To better capture global information, a pyramid approach is typically used, where features are progressively downsampled from the largest to the smallest resolution. The initial stages output the highest resolution, which contains abundant local detail information but lacks global information fusion. As the resolution decreases, global information is increasingly integrated into the lower-resolution features, providing an effective source of information for the subsequent decoder.
We use the outputs from the last four stages of ResNet18 as the inputs to our decoder. Given an input image with dimensions $H\times W \times 3$, the features output by each stage of ResNet18, denoted as $R1$ to $R4$, have channel dimensions of 64, 128, 256, and 512, respectively.
We can observe that as the stage number increases, the spatial dimensions of the features gradually decrease while the number of channels increases. Higher stages contain richer global information, whereas lower stages preserve more local details.

\subsection{Terrace Convolutional Decoder}\label{sec:tnet_decoder}
In segmentation models, the encoder typically utilizes a pretrained model on the ImageNet~\cite{5206848} dataset. 
Especially in the field of remote sensing image semantic segmentation, the decoder is critically important for achieving an organic fusion of global context and local details. 
TNet receives the outputs from the four stages of the backbone and processes them through its decoder.
As shown in Figure \ref{fig:overview}, the architecture of TNet can be broadly divided into two components: 
(1) a progressively enlarging upsampling structure resembling a terrace, and 
(2) a segmentation head responsible for regressing the final image segmentation results. The number of channels at each stage is determined by the outputs of the backbone. 

Here, we continue using TNet-R as an example. First, the features $R4$ are processed through a $3\times3$ transpose convolution kernel with stride 2, producing features $T1$ that match the size of $R3$. Since $R4$ contains rich global context, this step aims to effectively propagate the context, preparing for the subsequent fusion of local details. To enable multi-scale fusion, $R4$, $R3$, and $R2$ are each passed through a $3\times3$ convolution to produce features $C1$, $C2$, and $C3$, respectively. Then $T2$ is obtained by fusing $T1$ with $C1$, and $T3$ is derived from fusing $T2$ with $C2$. Finally, $T3$ is further fused with $C3$ (derived from $R1$) to incorporate the finest details before the segmentation head.

UNet's typical decoder performs skip-addition only after convolution, normalization, and ReLU; this strategy only amplifies features without any suppression, potentially introducing redundancy in remote sensing segmentation. \textbf{In contrast, TNet adds features immediately after convolution and batch normalization, then applies a ReLU6 activation before proceeding to the next stage}. This design effectively uses local detail to suppress excessive global context.
Compared to ReLU, ReLU6 prevents feature values from becoming excessively large, ensuring they remain manageable during subsequent propagation. 
The next step then focuses only on the global context corrected by local details, indirectly ensuring stability. 
Overall, each step in the terrace-like structure performs fusion and suppression of local details and global context within its scale, progressively refining the details until the final step predicts fine-grained details. 
Finally, the segmentation head (SegHead) performs pixel-level classification to produce the segmentation map, which is then upsampled using bilinear interpolation to match the resolution of the original image.

The architecture of TNet is simple yet effective, showcasing a step-by-step fusion process where low-resolution, high-semantic features are progressively integrated with high-resolution local details, resembling a terrace-like structure. Compared to pooling or upsampling operations, transpose convolution selectively learns global features and propagates them to the next level. Additionally, the interaction between $R$ and $C$ enables learning the distribution of local information, ultimately achieving a learnable fusion of global and local features.

In the upcoming experimental section, we will provide further validation through detailed experiments.

\subsection{Loss Function}
The loss function consists of cross-entropy loss $\mathcal{L}_{ce}$ and dice loss $\mathcal{L}_{dice}$. The cross-entropy loss $\mathcal{L}_{ce}$ is defined as:
\begin{equation}
   \mathcal{L}_{ce}=-\frac {1}{N}\sum_{n=1}^{N}\sum_{k=1}^{K}y_{k}^{n}\log\hat{y}_{k}^{n} .  
\end{equation}
where $N$ is the number of samples, $K$ is the number of classes, $\hat{y}_{k}^{n}$ is the predicted probability, and ${y}_{k}^{n}$ is the ground truth.
The dice loss $\mathcal{L}_{dice}$ is defined as :
\begin{equation}
    \mathcal{L}_{\text{dice}} = -\frac{2}{N} \sum_{n=1}^{N} \sum_{k=1}^{K} \frac{\hat{y}_k^n y_k^n} {\hat{y}_k^n} + y_k^n.
\end{equation} 
which enhances performance under class imbalance by emphasizing high-confidence predictions.
The total loss $\mathcal{L}_{total}$ is:
\begin{equation}
    \mathcal{L}_{total} = \mathcal{L}_{ce} + \mathcal{L}_{dice}.
\end{equation}
balancing pixel-level accuracy and region-level segmentation performance.

\section{Experiments and Analysis}\label{sec:exp}
In this section, we evaluate two instantiations of our network: TNet-R (ResNet18 backbone) and TNet-C (ConvNeXt-Base backbone). 
We first compare our method against state-of-the-art (SOTA) algorithms on multiple datasets. 
We next examine TNet's adaptability to different backbones on the Vaihingen dataset. 
Finally, we perform ablation studies to assess the impact of various design components on TNet's performance.

\begin{table*}[t]
\centering
\begin{adjustbox}{width=\textwidth}
\begin{tabular}{lccccccccc}
\toprule
\textbf{Method} & \textbf{Backbone} & \textbf{Imp. surf.} & \textbf{Building} & \textbf{Low. veg.} & \textbf{Tree} & \textbf{Car} & \textbf{mF1} & \textbf{OA} & \textbf{mIoU} \\
\midrule
UNet \cite{ronneberger2015u} & - & 96.29 & 94.75 &  82.87 & 88.99 & 84.04 & 89.39 & 90.21 & 81.25 \\
DANet \cite{fu2019dual}         & ResNet18          & 90.00              & 93.90             & 82.20              & 87.30         & 44.50        & 79.60        & 88.20        & 69.40         \\
ABCNet \cite{li2021abcnet}          & ResNet18          & 92.70              & 95.20             & 84.50              & 89.70         & 85.30        & 89.50        & 90.70        & 81.30         \\
BANet \cite{wang2021transformer}           & ResT-Lite         & 92.23              & 95.23             & 83.75              & 89.92         & 86.76        & 89.58        & 90.48        & 81.35         \\
Segmenter \cite{strudel2021segmenter}       & ViT-Tiny          & 89.80              & 93.00             & 81.20              & 88.90         & 67.60        & 84.10        & 88.10        & 73.60         \\
CMTFNet \cite{10247595}         & ResNet50          & 90.61              & 94.21             & 81.93              & 87.56         & 82.77        & 87.42        & 88.71        & 77.95         \\
UNetFormer \cite{wang2022unetformer}      & ResNet18          & 92.70              & 95.30             & 84.90              & \underline{90.60}         & 88.50        & 90.40        & 91.00        & 82.70         \\
AFENet \cite{10955240}          & ResNet18          & 96.90              & 95.72             & 85.07              & \textbf{90.64}         & 89.37        & 91.54        & 91.67        & 84.55         \\
PyramidMamba \cite{wang2024pyramidmamba}    & ResNet18          & 97.00              & \underline{96.10}             & \underline{85.50}              & 90.30         & 89.20        & 91.60        & 93.70        & 84.80         \\
UMFormer \cite{10969832}        & ResNet18          & 96.70              & 95.20             & 83.80              & 89.50         & 88.10        & 90.70        & 93.00        & 83.30         \\
D2LS \cite{d2ls}            & ConvNeXt-Base     & -                  & -                 & -                  & -             & -            & 91.90        & -            & -             \\
\midrule
TNet-R (Ours)          & ResNet18          & \underline{97.02}              & 96.05             & 85.46              & 90.50         & \textbf{90.65}        & \underline{91.93}        & \underline{93.68}       & \underline{85.35}         \\
TNet-C (Ours)          & ConvNeXt-Base     & \textbf{97.06}              & \textbf{96.30}             & \textbf{85.55}              & 90.49         & \underline{90.63}        & \textbf{92.01}        & \textbf{93.79}        & \textbf{85.48}         \\
\bottomrule
\end{tabular}
\end{adjustbox}
\caption{Quantitative comparison results on the Vaihingen testing set with state-of-the-art networks. The best value in each column is shown in bold, and the second-best value is underlined.}
\label{tab:vaihingen}
\end{table*}

\begin{table*}[t]
\centering
\begin{adjustbox}{width=\textwidth}
\begin{tabular}{lccccccccc}
\toprule
\textbf{Method} & \textbf{Backbone} & \textbf{Imp. surf.} & \textbf{Building} & \textbf{Low. veg.} & \textbf{Tree} & \textbf{Car} & \textbf{mF1} & \textbf{OA} & \textbf{mIoU} \\
\midrule
UNet \cite{ronneberger2015u} & - & 92.66 & 94.89 &  86.17 & 87.71 & 95.36 & 91.35 & 89.94 & 84.36 \\
DANet \cite{fu2019dual}           & ResNet18          & 91.00              & 95.60             & 86.10              & 87.60         & 84.30        & 88.90        & 89.10        & 80.30         \\
ABCNet \cite{li2021abcnet}          & ResNet18          & 93.50              & 96.90             & \underline{87.90}              & 89.10         & 95.80        & 92.70        & 91.30        & 86.50         \\
BANet \cite{wang2021transformer}           & ResT-Lite         & 93.34              & 96.66             & 87.37              & \underline{89.12}         & 95.99        & 92.50        & 91.06        & 86.25         \\
Segmenter \cite{strudel2021segmenter}       & ViT-Tiny          & 91.50              & 95.30             & 85.40              & 85.00         & 88.50        & 89.20        & 88.70        & 80.70         \\
CMTFNet \cite{10247595}         & ResNet50          & 92.12              & 96.41             & 86.43              & 87.26         & 92.41        & 90.93        & 89.89        & 83.57         \\
UNetFormer \cite{wang2022unetformer}      & ResNet18          & 93.60              & \underline{97.20}             & 87.70              & 88.90         & 96.50        & 92.80        & 91.30        & 86.80         \\
UMFormer \cite{10969832}        & ResNet18          & 93.70              & 96.40             & 86.70              & 87.80         & 95.50        & 92.00        & 90.90        & 85.50         \\
\midrule
TNet-R (Ours)          & ResNet18          & \underline{94.38}              & 96.89             & 87.61              & 89.03         & \underline{96.75}        & \underline{92.93}        & \underline{91.66}        & \underline{87.05}         \\
TNet-C (Ours)          & ConvNeXt-Base     & \textbf{94.80}              & \textbf{97.31}             & \textbf{88.36}              & \textbf{89.57}         & \textbf{97.12}        & \textbf{93.43}        & \textbf{92.22}        & \textbf{87.91}         \\
\bottomrule
\end{tabular}
\end{adjustbox}
\caption{Quantitative comparison results on the Potsdam testing set with state-of-the-art networks. The best value in each column is shown in bold, and the second-best value is underlined.}
\label{tab:potsdam}
\end{table*}

\begin{table*}[t]
\centering
\begin{adjustbox}{max width=\textwidth}
\begin{tabular}{lcccccccc}
\toprule
\textbf{Method} & \textbf{Background} & \textbf{Building} & \textbf{Road} & \textbf{Water} & \textbf{Barren} & \textbf{Forest} & \textbf{Agriculture} & \textbf{mIoU} \\
\midrule
UNet \cite{ronneberger2015u} &  43.38 & 53.53 &  53.52 & 76.03 & 17.75 & 44.74 & 58.33 & 49.61 \\
ABCNet \cite{li2021abcnet}          & 53.00              & \underline{62.18}             & 52.42         & 62.02          & \textbf{29.80}           & 41.92           & 47.27              & 49.80         \\
BANet \cite{wang2021transformer}            & \underline{53.94}               & 62.14             & 51.33         & 64.59          & 27.07           & 43.86           & 48.12              & 50.15         \\
Segmenter \cite{strudel2021segmenter}       & 38.00               & 50.70             & 48.70         & 77.40          & 13.30           & 43.50           & 58.20              & 47.10         \\
UNetFormer \cite{wang2022unetformer}      & 44.70               & 58.80             & 54.90         & 79.60          & 20.10           & 46.00           & 62.50              & 50.73         \\
CM-UNet \cite{liu2024cm}         & \textbf{54.56}               &\textbf{64.13}             &55.51         &68.06    &\underline{29.62}         &42.92             &50.42            &52.17 \\  
\midrule
TNet-R (Ours)          & 45.25               & 55.90             & \underline{56.60}         & \underline{81.01}          & 18.57           & \textbf{46.58}           & 61.41              & \underline{52.19}         \\
TNet-C (Ours)          & 46.59               & 60.48             & \textbf{58.82}         & \textbf{81.69}          & 21.12           & \underline{46.28}           & \textbf{63.72}              & \textbf{54.10}         \\
\bottomrule
\end{tabular}
\end{adjustbox}
\caption{Quantitative comparison results on the LoveDA online testing set with state-of-the-art networks. The best value in each column is shown in bold, and the second-best value is underlined.}
\label{tab:loveda}
\end{table*}
In this section, we compare our method with various state-of-the-art (SOTA) algorithms on mainstream remote sensing datasets. Additionally, we provide visualizations comparing our method with UNetFormer and UMFormer, both of which use ResNet18 as their backbone.

\begin{figure}[t]
\centering
\includegraphics[width=1\columnwidth]{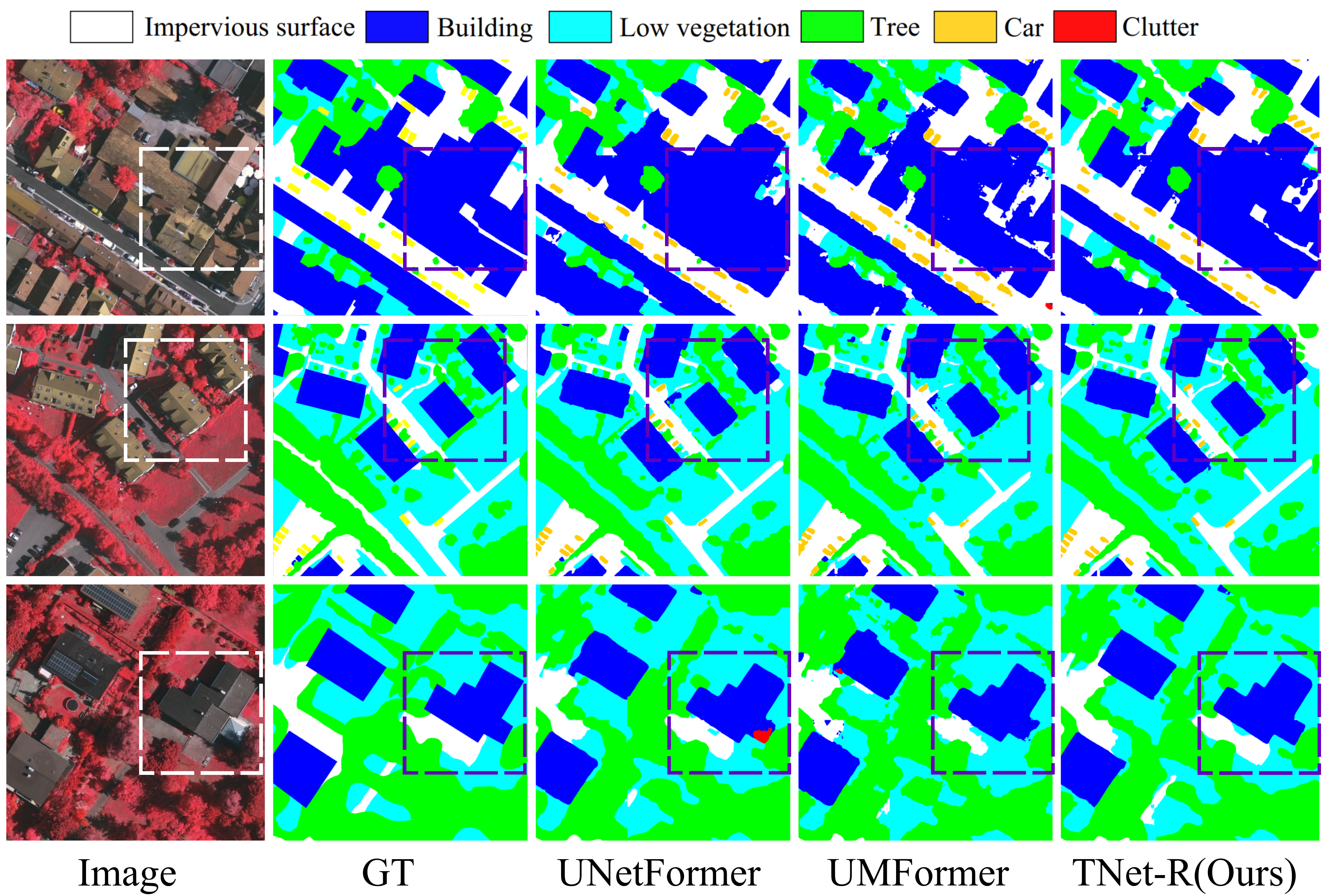} 
\caption{Visualization of the Vaihingen test set. Comparing our method with UNetFormer and UMFormer, our approach clearly identifies the edges of buildings even in occluded areas.}
\label{fig:val_vaihigen}
\end{figure}

\begin{figure}[t]
\centering
\includegraphics[width=1\columnwidth]{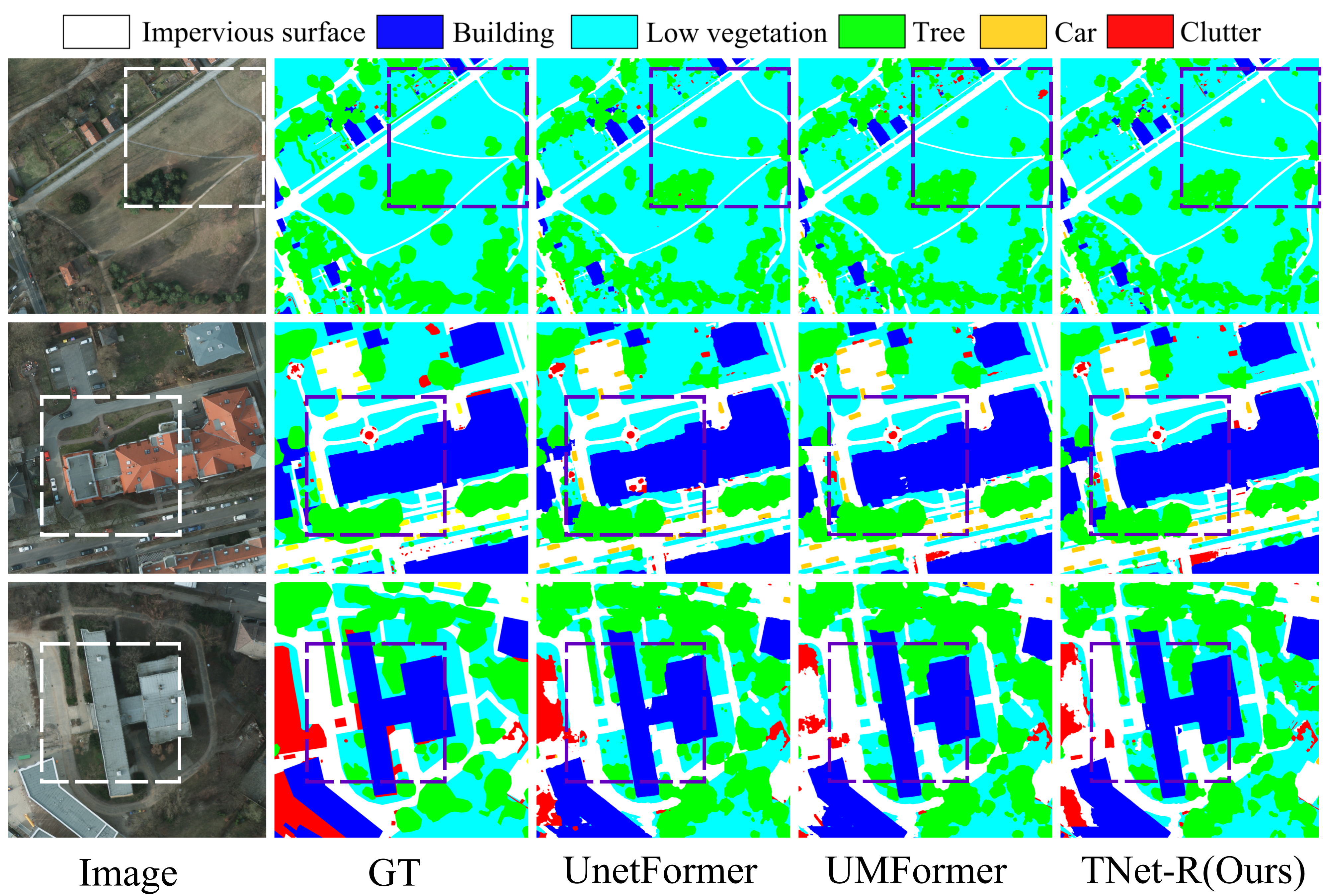} 
\caption{Visualization of the Potsdam test set. Comparing our method with UNetFormer and UMFormer, our approach demonstrates better control over misclassification.}
\label{fig:val_potsdam}
\end{figure}

\begin{figure}[t]
\centering
\includegraphics[width=1\columnwidth]{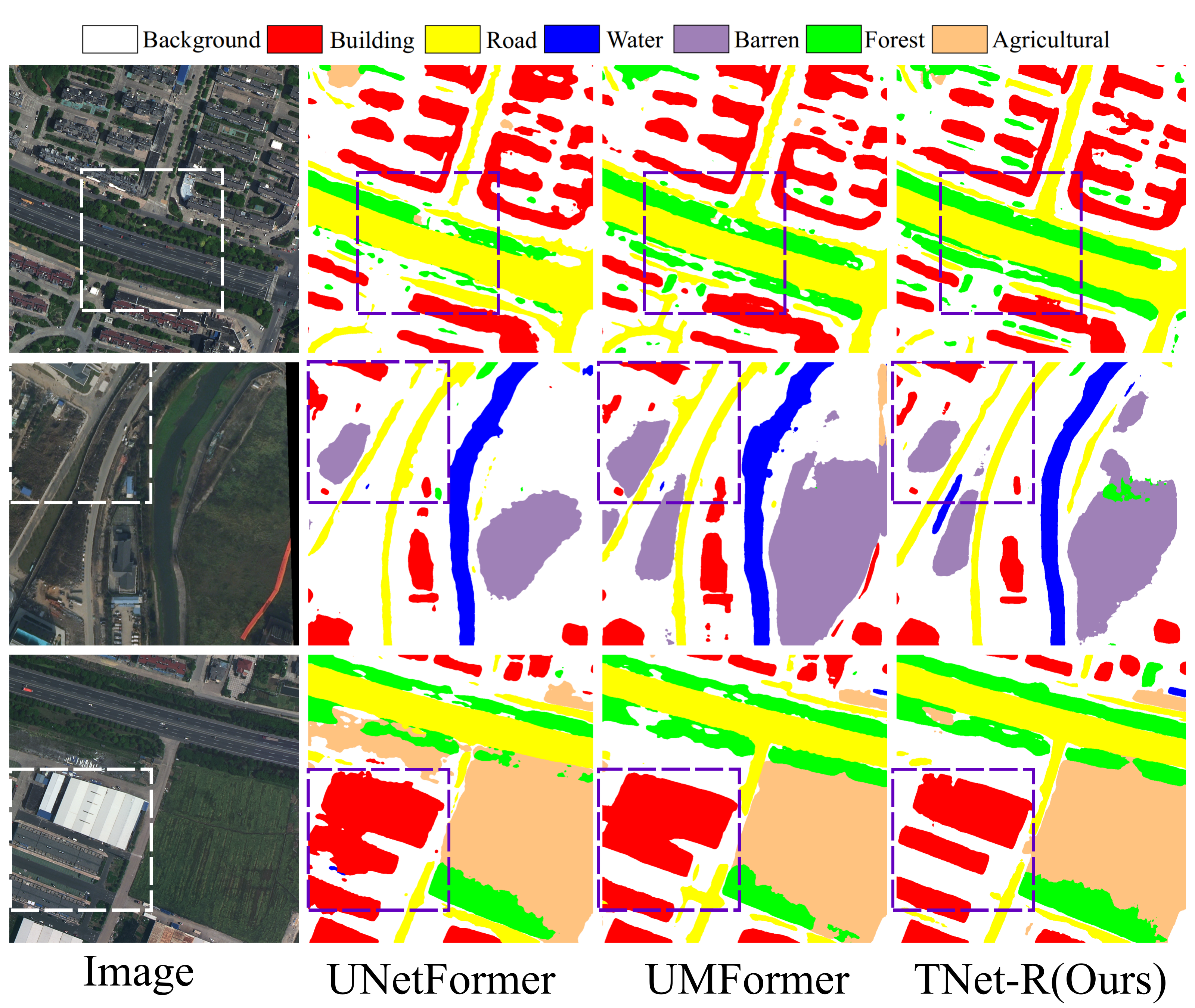} 
\caption{Visualization of the LoveDA test set. Comparing our method with UNetFormer and UMFormer, our approach achieves clearer recognition of road boundaries.}
\label{fig:val_loveda}
\end{figure}

\begin{figure}[t]
\centering
\includegraphics[width=1\columnwidth]{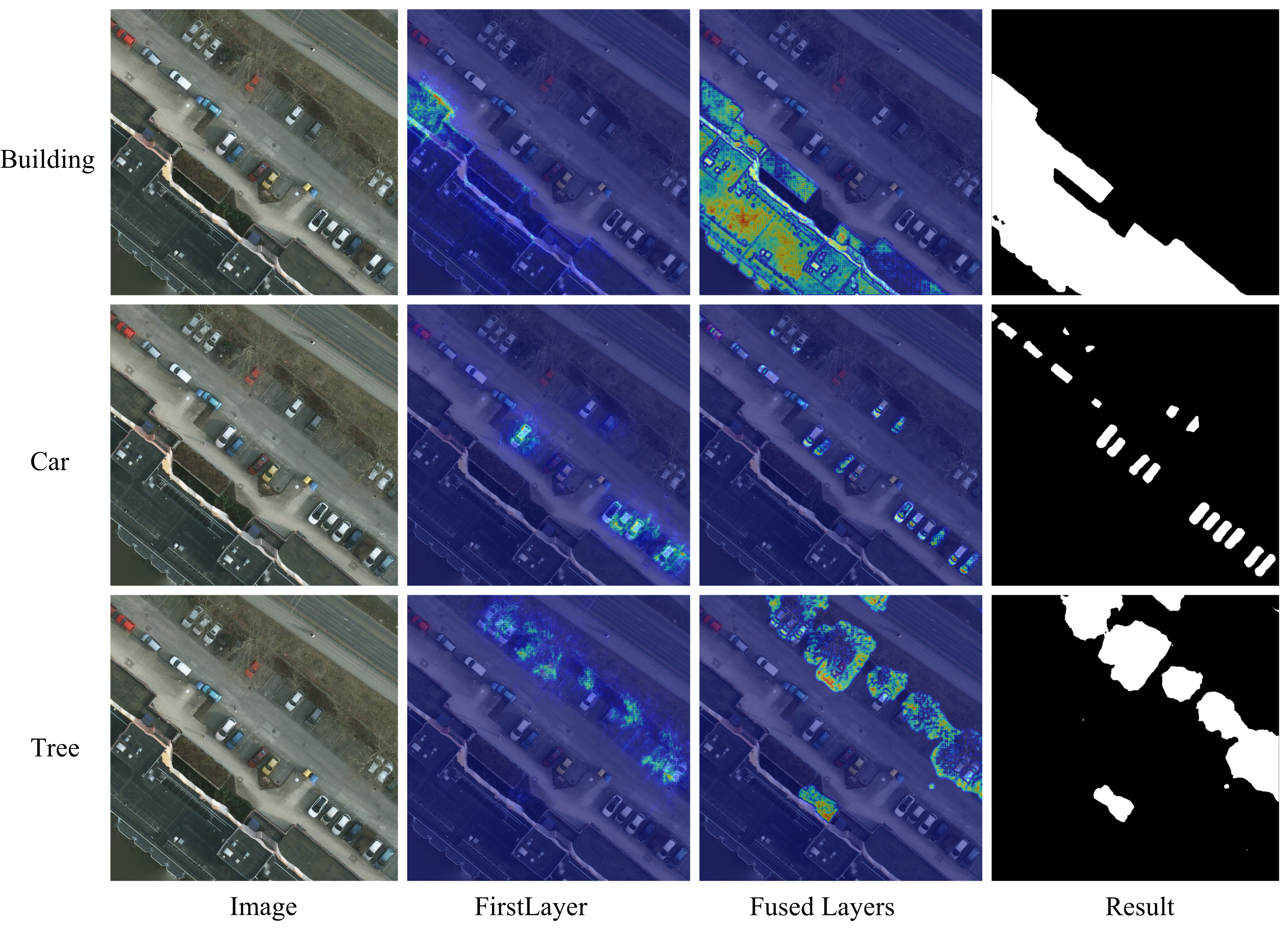} 
\caption{Visualization on layers in the Potsdam dataset. The First Layer corresponds to the features of R1, while the Fused Layer represents the features before entering the segmentation head.}
\label{fig:vis_final}
\end{figure}

\begin{table}[t]
\centering
\begin{adjustbox}{width=\columnwidth}
\begin{tabular}{lcccc}
\toprule
\textbf{Name} & \textbf{Backbone} & \textbf{mF1} & \textbf{OA} & \textbf{mIoU} \\
\midrule
TNet-R   & \multirow{2}{*}{ResNet18}        & \textbf{91.93}  & \textbf{93.68}  & \textbf{85.35}          \\
UNet-R   &                                  & 90.22  & 92.72  & 82.56 \\
\midrule
TNet-C   & \multirow{2}{*}{ConvNeXt-Base}   & \textbf{92.01}  & \textbf{93.80}  & \textbf{85.48}          \\
UNet-C   &                                  & 91.51  & 93.30  & 84.65 \\
\midrule
TNet-E   & \multirow{2}{*}{EfficientNet}    & \textbf{90.95}   & \textbf{93.36}  & \textbf{83.75}          \\
UNet-E   &                                  & 89.85   & 92.66   & 81.97 \\
\midrule
TNet-V   & \multirow{2}{*}{ViT-Tiny}        & \textbf{92.00}  &\textbf{93.77}   & \textbf{85.47}          \\
UNet-V   &                                  & 91.51  & 93.52   & 84.67 \\
\midrule
TNet-MV2 & \multirow{2}{*}{MobileNetV2}     & \textbf{91.19}   & \textbf{93.33}   & \textbf{84.11}          \\
UNet-MV2 &                                  & 89.61   & 92.37   & 81.57 \\
\midrule
TNet-E2  & \multirow{2}{*}{EfficientNetV2}  & \textbf{91.82}  & \textbf{93.65}  & \textbf{85.18}          \\
UNet-E2  &                                  & 90.78  & 92.95  & 83.46 \\ 
\bottomrule
\end{tabular}
\end{adjustbox}
\caption{Comparison of different backbone architectures with TNet and UNet in terms of mF1, OA, and mIoU on the Vaihingen Test Set. The best value in each column is shown in bold.}
\label{tab:backbone_comparison}
\end{table}

\begin{table}[t]
\centering
\begin{tabular}{lc}
\toprule
\textbf{Activation Layer} & \textbf{mIoU} \\
\midrule
ReLU6 & \textbf{85.35} \\
ReLU  & 85.10 \\
GELU  & \underline{85.11} \\
-     & 84.80 \\
\bottomrule
\end{tabular}
\caption{Impact of Different Activation Layers on TNet-R. Test on the Vaihingen Test Set. The best value in each column is shown in bold, and the second-best value is underlined.}
\label{tab:activation_layer}
\end{table}

\begin{figure}[t]
\centering
\includegraphics[width=1\columnwidth]{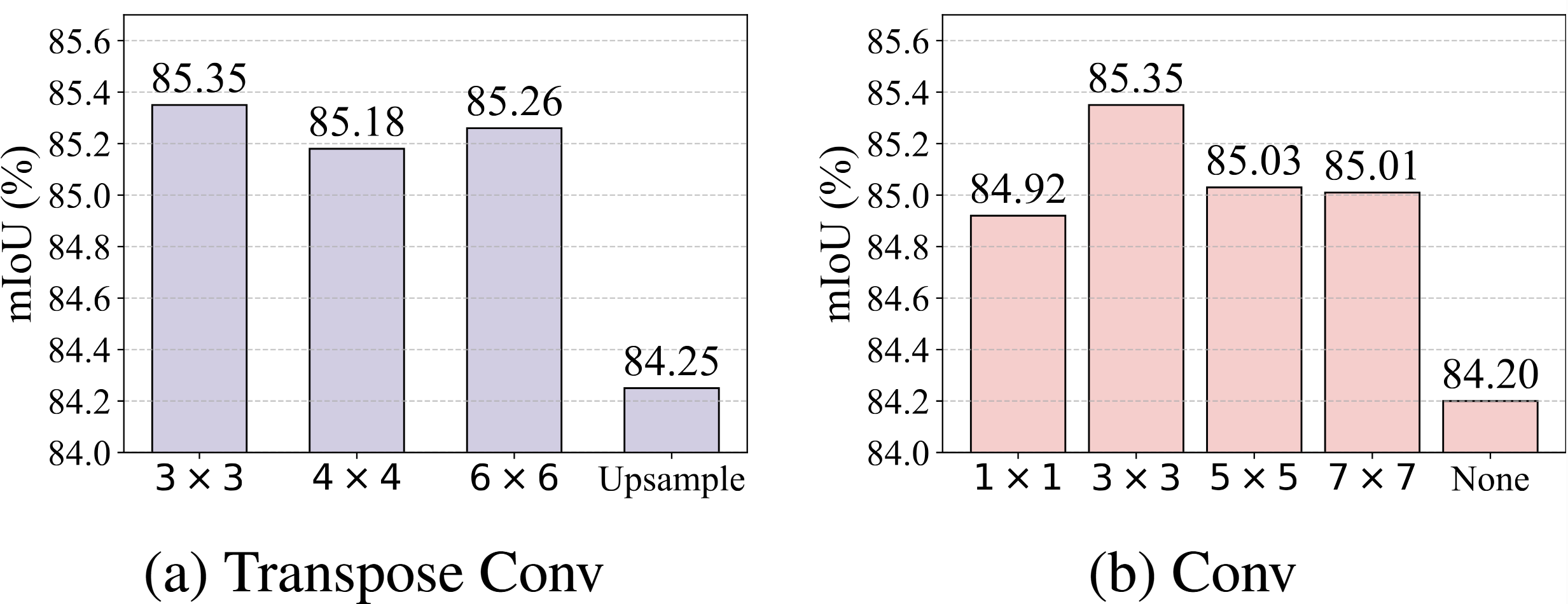} 
\caption{Impact of different transpose convolution and convolution kernel sizes on TNet-R. Test on the Vaihingen Test Set.}
\label{fig:deconv_kernel_size}
\end{figure}


\begin{table}[t]
\centering
\begin{adjustbox}{width=\columnwidth}
\begin{tabular}{lccccc}
\toprule
\textbf{Method} & \textbf{mIoU}$\uparrow$ & \textbf{FPS}$\uparrow$ & \textbf{Memory Usage}$\downarrow$ \\
\midrule
BANet                 & 81.35          & 229.44          & 558.45M          \\
ABCNet                & 81.30          & \underline{601.71}          & \underline{284.49M}          \\
UNetFormer  & 82.70          & 490.69          & 285.79M          \\
CMTFNet     & 77.95          & 227.00          & 1492.08M         \\
UMFormer    & \underline{83.30}& 67.14           & 724.22M          \\
TNet-R       & \textbf{85.35}          & \textbf{1037.15}         & \textbf{210.33M}          \\
\bottomrule
\end{tabular}
\end{adjustbox}
\caption{Comparison of different methods in terms of mIoU, FPS, and memory usage. Test on the Vaihingen Test Set. The best value in each column is shown in bold, and the second-best value is underlined.}
\label{tab:method_comparison}
\end{table}

\subsection{Datasets}
We conducted experiments on three widely used datasets: ISPRS Vaihingen, ISPRS Potsdam, and LoveDA [59]. These datasets serve as benchmarks in the field of remote sensing image semantic segmentation, encompassing diverse surface cover types, environmental conditions, and scene complexities. By leveraging these datasets, we ensured the broad applicability of our method and enabled comparisons with existing state-of-the-art approaches.

\textbf{\emph{LoveDA}} is a dataset comprising 5,987 high spatial resolution images annotated with seven semantic classes, a total of 166,768 labelled objects, and a Ground
Sampling Distance (GSD) of 0.3 m \cite{junjue_wang_2021_5706578}. Among these, 2,522 images are used for training, 1,669 for validation, and 1,796 are officially provided for testing. In our experiments, each image has a fixed size of $1024 \times 1024$ pixels.

\textbf{\emph{Vaihingen}} is a dataset consisting of 33 high spatial resolution images, with an average size of $2494 \times 2064$ pixels  and a Ground
Sampling Distance (GSD) of 9 cm \cite{ISPRS_Vaihingen_and_ISPRS_Potsdam_2024}. The dataset is annotated into six semantic
classes: impervious surfaces, buildings, low vegetation, trees,
cars, and clutter/background. In this study, we use 17 images (IDs 2, 4, 6, 8, 10, 12, 14, 16, 20, 22, 24, 27, 29, 31, 33, 35, and 38) for testing, while the remaining 16 images are used for training. In our experiments, each image is cropped into patches of size $1024 \times 1024$ pixels.

\textbf{\emph{Potsdam}} is a dataset consisting of 38 high spatial resolution images, each with a resolution of $6000 \times 6000$ pixels and a GSD of 5 cm \cite{ISPRS_Vaihingen_and_ISPRS_Potsdam_2024}. In this study, we use 15 images (IDs 2\_13, 2\_14, 3\_13, 3\_14, 4\_13, 4\_14, 4\_15, 5\_13, 5\_14, 5\_15, 6\_13, 6\_14, 6\_15, and 7\_13) for testing, while the remaining 23 images are used for training (excluding image 7\_10 due to erroneous annotations). In our experiments, each image is cropped into patches of size $1024 \times 1024$ pixels.

\subsection{Evaluation Metrics}
We followed the evaluation protocols of the comparative methods \cite{d2ls}\cite{hwang2024sfa}, including Overall Accuracy (OA), Intersection over Union (IoU), F1 Score, Mean IoU (mIoU), and Mean F1 (mF1). mIoU and mF1 represent the averages of IoU and F1 scores across all classes. For ease of reading, additional details are provided in the Appendix.

\subsection{Implementation Details}
All backbones used in this study are pretrained on the ImageNet dataset \cite{5206848} and sourced from the timm library \cite{rw2019timm}. The training process is conducted on an NVIDIA GeForce RTX 4090 GPU, with a batch size of 8, utilizing the AdamW \cite{loshchilov2017decoupled} optimiser with a one-cycle learning rate schedule. The peak learning rate is set to  
$6\times10^{-4}$, and the weight decay is configured to 0.001. Following the methodology outlined in \cite{wang2022unetformer}, for the Vaihingen, Potsdam, and LoveDA datasets, the images are randomly cropped into $512\times512$ patches.

During the training phase, several data augmentation techniques are employed to enhance model robustness. These include random scaling with factors $[0.5,0.75,1.0,1.25,1.5]$, random vertical flipping, random horizontal flipping, and random rotation. For the Vaihingen, Potsdam, and LoveDA datasets, the number of training epochs is set to 155, 100, and 120, respectively, with a batch size of 8 across all datasets. In the testing phase, multi-scale augmentations and random flipping are applied to further improve the model's performance and generalization capabilities.

\subsection{Comparison with State-of-the-art Methods}

\subsubsection{Comparisons With State-of-the-Art Methods on Vaihingen}
As shown in Table \ref{tab:vaihingen}, both TNet-C and TNet-R achieve the highest overall
performance on the Vaihingen test set across mF1, OA, and mIoU. For example, TNet-R exceeds the next-best method’s mIoU by 0.65\% (85.35\% vs. 84.70\% for PyramidMamba). 
Notably, our model handles
small objects (e.g., cars) especially well – it achieves the top IoU for the car class, indicating a clear
improvement in detecting tiny objects. 
Additionally, thanks to the organic fusion of global context and local details, as shown in Figure \ref{fig:val_vaihigen}, our method TNet-R demonstrates a clear advantage over UNetFormer and UMFormer. 
Specifically, it can accurately distinguish between building types, especially at the edges of buildings, achieving results that are closer to the ground-truth labels. 
\subsubsection{Comparisons With State-of-the-Art Methods on Potsdam}
As shown in Table \ref{tab:potsdam}, our TNet-C and TNet-R continue to achieve the best performance. Notably, they rank first in both large-scale objects, such as buildings, and small-scale objects, such as cars. This demonstrates that our TNet effectively integrates global large-scale information with local fine details. In Figure \ref{fig:val_potsdam}, it can be observed that our method, TNet-R, clearly identifies the boundaries and extents of buildings while effectively controlling misclassified areas. Additionally, in the segmentation of "Clutter," the misclassification rate is relatively low. Moreover, in complex scenes where "Trees" and "Cars" are interwoven, TNet-R is the only model capable of accurately identifying "Cars". This highlights TNet's powerful ability to encode global context and fuse local details.
\subsubsection{Comparisons With State-of-the-Art Methods on LoveDA}
As shown in Table \ref{tab:loveda}, our TNet-C and TNet-R achieve the top two rankings in the mIoU metric. Furthermore, they exhibit outstanding performance in categories such as water and road, which heavily rely on both global and local information.
From Figure \ref{fig:val_loveda}, it is evident that TNet not only handles the edges of buildings in a manner more consistent with the image's representation but also captures finer details for trees along the sides of roads. Additionally, TNet demonstrates superior road recognition capabilities compared to other algorithms, maintaining the continuity of roads without breaking them, thereby preserving their integrity. Compared to other methods, especially in scenarios requiring global contextual information, such as "Roads" and "Forests", TNet showcases powerful performance.

\subsection{Ablation Studies and Analysis}

\subsubsection{Different backbones with TNet and UNet}
As shown in Table \ref{tab:backbone_comparison}, we not only compare the performance of TNet with different backbone architectures but also analyze the performance of using a UNet decoder with the same backbones.

It can be observed that, on average, TNet achieves an improvement of approximately 1\% mIoU compared to UNet. Furthermore, TNet demonstrates superior fusion capabilities when paired with backbones such as ResNet18, MobileNetV2, EfficientNetV2, and EfficientNet.

\subsubsection{Different activation layers on TNet}
As shown in Table \ref{tab:activation_layer}, we investigated the impact of different activation functions on TNet by removing or replacing them and recording the mIoU results on the Vaihingen dataset. From the observations, ReLU6 yielded the highest mIoU (85.35\%).
This indicates ReLU6 is the most effective choice for TNet among the tested options.

\subsubsection{Different convolution kernel sizes on TNet-R}
As shown in Figure \ref{fig:deconv_kernel_size}, different convolution kernel sizes have a significant impact on TNet.For the decoder's transposed convolution (upsampling) layer, we experimented with a non-learned upsample operation in place of transpose convolution. The result (Figure \ref{fig:deconv_kernel_size} (a): 84.25\% mIoU with simple upsampling vs 85.35\% with $3\times3$ Transpose convolution ) demonstrates the value of a learnable convolution for upsampling. Furthermore, among transpose convolution kernel sizes, $3\times3$ produced the best result; larger kernels ($4\times4$ or $6\times6$) did not improve performance.

Similarly, for the convolution layer before addition, Figure \ref{fig:deconv_kernel_size} (b) shows that a $3\times3$ kernel yields the highest mIoU. 
Using a $1\times1$ or larger kernel, such as $5\times5$ or $7\times7$, slightly lowers accuracy, and omitting this convolution altogether causes a notable performance drop.

\subsubsection{Visualizations}
As shown in Figure \ref{fig:vis_final}, TNet effectively integrates features from different scales through multi-scale feature fusion, ultimately regressing information that encompasses both global context and local details. This is particularly evident in regions where Car and Tree intersect, which require a deeper understanding of contextual relationships. Additional visualization results for more layers are provided in the Appendix.

\subsubsection{Complexity Analysis}
Table \ref{tab:method_comparison} presents a comparison of resource consumption between TNet-R and various methods under the same input image size of $3\times512\times512$. Benefiting from its simple Conv-based architecture, TNet-R achieves an exceptionally high inference speed (1037 FPS on an RTX 4090) while maintaining top accuracy. This highlights that our network delivers state-of-the-art accuracy with significantly lower computational cost than its counterparts.

\section{Conclusion}
In this paper, we have presented TNet, a terrace-shaped convolutional decoder,
and demonstrated its effectiveness across multiple benchmarks (Vaihingen, Potsdam, LoveDA), including
state-of-the-art results with a ResNet18 backbone. We also showed TNet’s flexibility by integrating
various backbones without loss of performance. For future work, we plan to refine our network to further
boost accuracy in certain challenging regions of the image (such as boundaries or small-object areas). We
will also explore hybrid architectures that combine CNN-based decoders with other powerful models
(e.g., Transformers) to advance remote sensing image segmentation.

\bibliography{aaai2026}


\end{document}